\pdfoutput=1
\documentclass[11pt]{article}

\usepackage[]{acl}

\usepackage{times}
\usepackage{latexsym}
\usepackage{multicol}
\usepackage{multirow}

\usepackage[T1]{fontenc}
\usepackage[utf8]{inputenc}

\usepackage{booktabs}

\usepackage{microtype}

\usepackage{inconsolata}

\usepackage{graphicx}

\usepackage{enumitem}

\title{Refining Financial Consumer Complaints through Multi-Scale Model Interaction}

\author{
 \textbf{Bo-Wei Chen,\textsuperscript{1}}
 \textbf{An-Zi Yen,\textsuperscript{1}}
 \textbf{Chung-Chi Chen\textsuperscript{2}}
\\
\\
  \textsuperscript{1}Department of Computer Science, National Yang Ming Chiao Tung University, Taiwan\\
   \textsuperscript{2}National Institute of Advanced Industrial Science and Technology, Japan \\
  \texttt{h7a4n1k.cs12@nycu.edu.tw, azyen@nycu.edu.tw, c.c.chen@acm.org}\\
}

\begin{document}
\maketitle
\begin{abstract}
Legal writing demands clarity, formality, and domain-specific precision—qualities often lacking in documents authored by individuals without legal training. To bridge this gap, this paper explores the task of legal text refinement that transforms informal, conversational inputs into persuasive legal arguments. We introduce FinDR, a Chinese dataset of financial dispute records, annotated with official judgments on claim reasonableness. Our proposed method, Multi-Scale Model Interaction (MSMI), leverages a lightweight classifier to evaluate outputs and guide iterative refinement by Large Language Models (LLMs). Experimental results demonstrate that MSMI significantly outperforms single-pass prompting strategies. Additionally, we validate the generalizability of MSMI on several short-text benchmarks, showing improved adversarial robustness. Our findings reveal the potential of multi-model collaboration for enhancing legal document generation and broader text refinement tasks. 
\end{abstract}

\section{Introduction}

People compose a variety of texts in their daily lives, and they often need expert advice for formal documents. In the legal domain, legal writing requires precise presentation of facts, strong argumentation, and the usage of specialized vocabulary \cite{neumann2021legal}, which can be challenging for non-professionals. People without a legal background may struggle to express their concerns using legal language, resulting in more conversational texts compared to those written by professionals such as lawyers. This can lead to disadvantages such as applications being more likely to be rejected or claims less likely to be adopted. Moreover, hiring legal experts to assist with such documents can be costly, placing a financial burden on individuals and small businesses.
To address this issue, automated tools and methodologies that assist in generating legally sound documents can significantly reduce the need for human intervention, thus lowering costs and improving accessibility to legal expertise.

\begin{figure}[t]
  \includegraphics[width=\columnwidth]{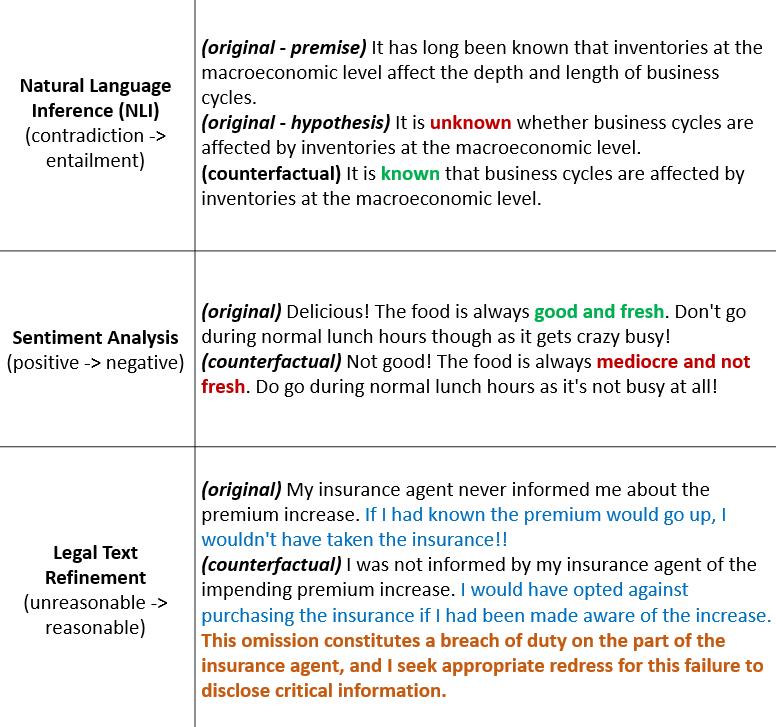}
  \caption{Comparison of generated counterfactual texts between our purposed task and the other classification tasks. The example of the legal text refinement task is a plaintiff’s claim in an insurance dispute, sampled from our dataset.}
  \label{fig:dataset}
\end{figure}

\begin{table*}[t]
    \centering
    \small
    \begin{tabular}{lcccccc}
    \toprule
    \multicolumn{1}{c}{\textbf{Dataset}} & \textbf{Samples} & \textbf{Avg. Words} & \textbf{Labels} \\
    \hline
    IMDB \cite{maas-etal-2011-learning} & 50,000 & 234 & positive, negative \\
    Yelp \cite{Zhang2015CharacterlevelCN} & 598,000 & 133 & positive, negative \\
    AGNews \cite{Zhang2015CharacterlevelCN} & 127,600 & 37 & world, sports, business, sci/tech \\
    MNLI \cite{williams-etal-2018-broad} & 431,992 & 20 / 10 & entailment, contradiction, neutral \\
    DBPedia \cite{10.1007/978-3-540-76298-0_52} & 630,000 & 46 & 14 labels \\
    \hline
    FinDR (Proposed) & 13,132 & 598 & reasonable, unreasonable \\
    \bottomrule
    \end{tabular}
    \caption{Comparing the statistics of FinDR with various popular counterfactual generation datasets. For MNLI, average words is denoted by (premise / hypothesis).}
    \label{tab:exp-event-type}
\end{table*}

Recent advancements in Large Language Models (LLMs) have demonstrated strong capabilities in counterfactual generation \cite{dixit-etal-2022-core, chen-etal-2023-disco, nguyen2024llmsgeneratingevaluatingcounterfactuals}. These models can produce fluent counterfactual texts that rival or even surpass human-written content. However, most existing work has focused on short texts, such as those found in Natural Language Inference (NLI) or sentiment analysis tasks, as shown in Table~\ref{tab:exp-event-type}. These tasks often involve simple modifications like negation or word replacement to change the label of a sentence.
In contrast, generating counterfactuals for longer and more complex documents, such as legal texts, presents unique challenges. Legal documents often involve intricate arguments from plaintiffs, defendants, and judges. Refining or rewriting such documents requires a deep understanding of the case and modifications at the sentence or paragraph level, rather than simple word-level changes.

To explore this direction, we introduce FinDR,\footnote{\url{https://github.com/NYCU-NLP-Lab}} a Chinese legal document dataset composed of \textbf{Fin}ancial \textbf{D}ispute \textbf{R}ecords from real-world cases. FinDR is designed to push the boundaries of counterfactual generation by requiring models to generate more persuasive and formally structured legal claims. Figure~\ref{fig:dataset} shows how counterfactual text generation in FinDR differs from that in typical tasks.
In NLI and sentiment analysis, counterfactual rewriting focuses on semantic or sentiment polarity shifts.
In contrast, legal text rewriting demands more sophisticated domain-specific knowledge and rigorous logical reasoning. 
It requires not only adherence to the original factual context but also transforms informal or unreasonable statements into formal claims with legal validity.

In this work, we focus on the task as legal text refinement, where the goal is to transform colloquial writing into polished, formal legal arguments. A central question in refinement is: \textit{``How can we determine whether one version of a text is better than another?''} This is a challenging question, as writing quality can be subjective and context-dependent.
To address this, we propose using well-trained classifiers as discriminating models to evaluate the refined output. These models serve as proxies for legal or linguistic judgment, helping to quantify improvements in quality, relevance, and persuasiveness.

%To sum up, the contributions of our work are as follows:
%\begin{enumerate}[nolistsep]
% \item We propose the task of legal text refinement to explore the transformation of colloquial language into professional legal writing.
% \item We introduce FinDR, a challenging Chinese legal document dataset that demands nuanced understanding and sentence-level editing.
% \item We explore a novel approach that leverages counterfactual generation with LLMs, using discriminator feedback to guide and evaluate refinement quality.
%\end{enumerate}

\section{Related Work}
%\textbf{Text Refinement on Large Language Models.}
To obtain better quality of output texts, several works have explored the ``self-refining'' and ``self-correcting'' abilities of LLMs, where they revise their responses based on internal feedback to their previous outputs.
\citet{madaan2023selfrefineiterativerefinementselffeedback} proposed an iterative refinement framework that can obtain better output texts without the need of additional training or fine-tuning.
\citet{shinn2023reflexionlanguageagentsverbal} takes a step further, leveraging LLMs to convert feedback from the task environment, usually in binary or scalar form, to a verbal text summary, which is then provided as additional context for generation.

However, a few studies have pointed out that these self-correction schemes are not reliable.
\citet{huang2024largelanguagemodelsselfcorrect} showed that without any external or human feedback, self-correcting models are not confident about their response, often revising a correct answer to an incorrect one, thus causing a performance drop.
\citet{xu-etal-2024-pride} formalized the evaluation of self-bias of LLMs, and demonstrated that when an evaluation step is included in a self-refine pipeline, the self-bias is amplified, favoring its own generation and deviating from human evaluation.
Therefore, we leverage external feedback from trained classifiers or other LLMs to prevent self-bias and potential performance decrease.

%\textbf{Counterfactual Generation using Large Language Models.} \cite{nguyen2024llmsgeneratingevaluatingcounterfactuals} 

% 介紹 CF generation
Counterfactual data are adversarial inputs designed to flip a model's prediction with subtle changes to the original inputs \cite{kaushik2020learningdifferencemakesdifference}.
In NLP, this involves word perturbations, text additions and deletions, or even the whole structure.
While counterfactual generation is often discussed as a method of data augmentation~\citep{madaan2021generatecounterfactualscontrolledcounterfactual, paranjape-etal-2022-retrieval}, in this paper, we propose leveraging counterfactual generation as a means of text refinement.
For example, in binary classification, given an accurate discriminating model $M$ and a text $t$ that is labelled ``negative'' (denoting ``bad text''), the refinement goal is to generate a counterfactual text $t_c$ that causes $M$ to flip its prediction to positive (denoting ``good text''), we call $t_c$ a refinement of $t$.
% 好像缺了一點甚麼，方法的好處？

\section{Dataset}
%\subsection{Data Collection and Preprocessing}
We crawled financial dispute records that are public available from Taiwan Financial Ombudsman Institution.\footnote{The Financial Ombudsman Institution is a government organization funded by Taiwanese government that aims to fairly, reasonably, and effectively resolve financial disputes between consumers and enterprises.}
Originally, these records were legal documents in pdf format.
We extract the content and perform data cleaning to remove texts such as tables, page numbers or links.
We collected a total of 13,132 records, with dates ranging from February 2012 to March 2023.
Each record consists of three primary columns: the plaintiff's claim, the defendant's rebuttal, and the judgment made by the officials, and a label.

There are 6 possible labels attached to a record, with the most frequent two labels being ``reasonable'' and ``unreasonable''.
A record is labeled ``reasonable'' if the officials find the plaintiff's claim to be reasonable, and vice versa.
The remaining 3 labels are ``some reasonable'', ``some unreasonable'', and ``some not applicable'', indicating that certain parts of the claim are accepted, not accepted, and irrelevant, respectively.
For the purpose of counterfactual generation, we use a modified version of the dataset where labels are remapped by the following rule: if the label contains ``reasonable'' or ``some reasonable'', then it will be remapped to ``reasonable'', otherwise ``unreasonable''.

%\subsection{Data Statistics}
% TODO: Table

\section{Method}
We divide the refinement task into two phases: evaluation and refinement. These tasks are assigned to models of different scales—specifically, a small pre-trained language model (PLM)
%, such as BERT, 
is employed to evaluate whether an input is reasonable or unreasonable. Based on the evaluation results provided by the PLM, an LLM performs the refinement. Through several rounds of interaction, the objective is to enable the PLM to judge inputs that were initially considered unreasonable as reasonable. We name this approach Multi-Scale Model Interaction (MSMI).

\subsection{Discriminating Model for Evaluation}
We fine-tune a binary classification model to predict whether a given legal text (typically, a plaintiff's claim) is reasonable or not. The model is trained using our labeled FinDR dataset, where the labels correspond to the judgment outcomes: ``reasonable'' vs. ``unreasonable''. 
We employed the BERT-base model~\citep{devlin2019bert} as the PLM and selected the best performing checkpoint on the validation set as the discriminating model $M$. During refinement, this model provides feedback to the generator, indicating whether a new version of the input text flips the classification outcome.

\subsection{LLM-based Refinement via Counterfactual Generation}
To improve an ``unreasonable'' input $t$ into a ``reasonable'' version $t_c$, we employ an LLM as a generator to perform the rewriting. The LLM is prompted with the original text, alongside context and instructions.
To achieve more effective refinements, we perform multiple rounds of generation and evaluation. After each round, the LLM receives feedback indicating whether the new version was accepted by the classifier. If not, the model is encouraged to revise again, using reinforcement or rejection cues based on the classifier’s confidence.

\subsection{Iterative Loop and Early Stopping}
We adopt an iterative loop similar to the self-refinement setting~\citep{madaan2023selfrefineiterativerefinementselffeedback}. However, instead of using internal self-evaluation, we use external signals from $M$. In each iteration:
\begin{enumerate}[nolistsep]
  \item The LLM generates a refined version $t_c$ of $t$.
  \item The classifier evaluates $t_c$ and returns a probability score.
  \item If the score crosses a threshold, we stop and accept the result.
  \item Otherwise, we provide feedback and loop again.
\end{enumerate}

\section{Experiments}
We use the success rate to measure how effectively our method transforms documents from unreasonable to reasonable refinements. Additionally, we compute the cosine similarity between the original and refined document embeddings to assess the degree of semantic change. A high cosine similarity indicates that the refinement preserves the original meaning while making the text more reasonable, whereas a low similarity may suggest substantial rewriting or a shift in semantic content.

First, regarding the classifier, we found that BERT was able to achieve an accuracy of 80.5\%, and thus we adopted BERT as the scoring model. As for the LLMs, we used GPT-3.5~\citep{ouyang2022training}, GPT-4o~\citep{hurst2024gpt}, and LLaMA-3~\citep{grattafiori2024llama} for refinement. We also used the approach of simply prompting the LLM without a classifier as the baseline to evaluate the effectiveness of the proposed MSMI method.

Table~\ref{tab: Experimental results in FinDR} presents the experimental results. First,  regardless of LLM is used, the MSMI strategy consistently improves the success rate, demonstrating the generalizability of our method. Second, based on the cosine similarity results, MSMI introduces slightly more content variation, meaning the final text tends to undergo greater changes after multiple rounds of modification. Third, we found another advantage of MSMI—it may eliminate the need to select a specific LLM. From the results of the original prompts, there is a noticeable performance gap among different LLMs. However, after applying MSMI, even GPT-3.5 achieves performance comparable to GPT-4o and LLaMA-3.

In sum, the experimental findings validate the effectiveness and robustness of the MSMI method. Not only does it consistently improve performance across different LLMs, but it also helps mitigate disparities between models of varying capabilities. This makes MSMI a practical solution in real-world scenarios where access to the best-performing LLMs may be limited due to cost or availability. These results collectively highlight MSMI as an ideal enhancement strategy for leveraging LLMs in iterative refinement tasks.

\begin{table}[t]
  \centering
  \resizebox{\columnwidth}{!}{
    \begin{tabular}{ll|rr}
    \multicolumn{1}{c}{Strategy} & \multicolumn{1}{c|}{LLM} & \multicolumn{1}{c}{\textbf{Success Rate}~($\uparrow$)} & \multicolumn{1}{c}{\textbf{Cosine Sim.}~($\uparrow$)} \\
    \hline
    \multirow{3}[2]{*}{Prompt} & GPT-3.5 & 8.70\% & 0.943 \\
    & GPT-4o & 16.00\% & 0.974 \\          
          & LLaMA-3-70B & 13.10\% & 0.957 \\
    \hline
    \multirow{3}[1]{*}{MSMI (Proposed)} & GPT-3.5 & 25.00\% & 0.943 \\
    & GPT-4o & 26.10\% & 0.962 \\
          & LLaMA-3-70B & 26.10\% & 0.946 \\
    \end{tabular}
    }
  \caption{Experimental results in FinDR.}
  \label{tab: Experimental results in FinDR}%
\end{table}%

\section{Further Exploration}

As previously mentioned, most earlier tasks focused on short-text refinement. In this section, we evaluate the performance of the proposed MSMI model under this specific task. Experiments are conducted using the five benchmarks listed in Table~\ref{tab:exp-event-type}. To enable comparison with prior studies, we adopt their methodology by using ``Adversarial Accuracy'' as the evaluation metric.

Adversarial Accuracy measures a model’s robustness against adversarial attacks by evaluating its prediction accuracy on adversarially perturbed inputs. Specifically, it reflects the percentage of adversarial examples on which the model correctly predicts the label. A lower Adversarial Accuracy indicates a more effective attack, as the model is more easily misled by perturbations. Conversely, a higher Adversarial Accuracy suggests that the model is more resilient and better able to withstand adversarial manipulations. Therefore, in our experiments, Adversarial Accuracy is considered a ``the lower, the better'' metric.

% Table~\ref{tab:short-text refinement tasks} presents the experimental results.
As shown in Table~\ref{tab:short-text refinement tasks}, in four out of the five datasets, MSMI outperforms the baseline approach of directly using LLMs. This indicates that, even in short-text scenarios, a single-pass refinement by an LLM may be insufficient. From the perspective of cosine similarity, multiple rounds of refinement tend to result in more substantial content changes—suggesting that the model is not merely inserting ``not'' or altering a single word, but may be rewriting the text to a greater extent after several iterations. In the context of short texts, this leads to more noticeable differences in cosine similarity.
Combining results from both long-text (legal) and short-text experiments, we demonstrate that MSMI is a simple yet effective approach that enhances both refinement quality and adversarial robustness.
% Combining the findings from both legal document (long-text) and short-text experiments, we demonstrate a simple yet efficient method. The use of MSMI contributes to improved performance in both refinement quality and robustness in adversarial attacks.

\begin{table}[t]
  \centering
    \resizebox{\columnwidth}{!}{
    \begin{tabular}{c|lrr}
    Dataset & \multicolumn{1}{c}{\textbf{Method}} & \textbf{Adv Acc.} ($\downarrow$) & \textbf{Cosine Sim.} ($\uparrow$) \\
    \hline
    \multirow{3}[2]{*}{MNLI} & GPT-3.5 & 18.6  & \textbf{0.96} \\
          & LLaMA-3-70B & 27.5  & \textbf{0.96} \\
          & MSMI (Proposed) & \textbf{4.7} & 0.80 \\
    \hline
    \multirow{3}[2]{*}{AG News} & GPT-3.5 & 12.7  & 0.86 \\
          & LLaMA-3-70B & 4.0   & \textbf{0.89} \\
          & MSMI (Proposed) & \textbf{3.5} & 0.80 \\
    \hline
    \multirow{3}[2]{*}{Yelp} & GPT-3.5 & 33.0  & 0.89 \\
          & LLaMA-3-70B & \textbf{2.8} & 0.89 \\
          & MSMI (Proposed) & 4.4   & \textbf{0.94} \\
    \hline
    \multirow{3}[2]{*}{IMDB} & GPT-3.5 & 10.2  & 0.90 \\
          & LLaMA-3-70B & 4.6   & 0.86 \\
          & MSMI (Proposed) & \textbf{1.8} & \textbf{0.96} \\
    \hline
    \multirow{3}[1]{*}{DBPedia} & GPT-3.5 & 36.5  & \textbf{0.90} \\
          & LLaMA-3-70B & 9.9   & 0.87 \\
          & MSMI (Proposed) & \textbf{7.1} & 0.80 \\
    \end{tabular}%
    }
  \caption{Experimental results in short-text refinement tasks.}
  \label{tab:short-text refinement tasks}%
\end{table}%

\section{Conclusion}
In this work, we tackle the challenge of transforming informal legal claims into refined, persuasive documents using large language models. We introduce the FinDR dataset, a collection of real-world Chinese financial dispute records, which serves as a benchmark for evaluating legal text refinement. Our proposed MSMI framework demonstrates clear advantages over single-step generation, consistently improving classification success rates while maintaining semantic coherence. Moreover, MSMI proves effective across both long and short-text scenarios, reducing performance gaps among different LLMs and enhancing adversarial robustness. These results suggest that collaborative model architectures, where small models offer targeted feedback to larger generators, hold great promise for advancing automated writing tools, especially in high-stakes domains like law.

%\section*{Acknowledgments}

\bibliography{custom}

\begin{thebibliography}{19}
\providecommand{\natexlab}[1]{#1}

\bibitem[{Auer et~al.(2007)Auer, Bizer, Kobilarov, Lehmann, Cyganiak, and Ives}]{10.1007/978-3-540-76298-0_52}
S{\"o}ren Auer, Christian Bizer, Georgi Kobilarov, Jens Lehmann, Richard Cyganiak, and Zachary Ives. 2007.
\newblock Dbpedia: A nucleus for a web of open data.
\newblock In \emph{The Semantic Web}, pages 722--735, Berlin, Heidelberg. Springer Berlin Heidelberg.

\bibitem[{Chen et~al.(2023)Chen, Gao, Bosselut, Sabharwal, and Richardson}]{chen-etal-2023-disco}
Zeming Chen, Qiyue Gao, Antoine Bosselut, Ashish Sabharwal, and Kyle Richardson. 2023.
\newblock \href {https://doi.org/10.18653/v1/2023.acl-long.302} {{DISCO}: Distilling counterfactuals with large language models}.
\newblock In \emph{Proceedings of the 61st Annual Meeting of the Association for Computational Linguistics (Volume 1: Long Papers)}, pages 5514--5528, Toronto, Canada. Association for Computational Linguistics.

\bibitem[{Devlin et~al.(2019)Devlin, Chang, Lee, and Toutanova}]{devlin2019bert}
Jacob Devlin, Ming-Wei Chang, Kenton Lee, and Kristina Toutanova. 2019.
\newblock Bert: Pre-training of deep bidirectional transformers for language understanding.
\newblock In \emph{Proceedings of the 2019 conference of the North American chapter of the association for computational linguistics: human language technologies, volume 1 (long and short papers)}, pages 4171--4186.

\bibitem[{Dixit et~al.(2022)Dixit, Paranjape, Hajishirzi, and Zettlemoyer}]{dixit-etal-2022-core}
Tanay Dixit, Bhargavi Paranjape, Hannaneh Hajishirzi, and Luke Zettlemoyer. 2022.
\newblock \href {https://doi.org/10.18653/v1/2022.findings-emnlp.216} {{CORE}: A retrieve-then-edit framework for counterfactual data generation}.
\newblock In \emph{Findings of the Association for Computational Linguistics: EMNLP 2022}, pages 2964--2984, Abu Dhabi, United Arab Emirates. Association for Computational Linguistics.

\bibitem[{Grattafiori et~al.(2024)Grattafiori, Dubey, Jauhri, Pandey, Kadian, Al-Dahle, Letman, Mathur, Schelten, Vaughan et~al.}]{grattafiori2024llama}
Aaron Grattafiori, Abhimanyu Dubey, Abhinav Jauhri, Abhinav Pandey, Abhishek Kadian, Ahmad Al-Dahle, Aiesha Letman, Akhil Mathur, Alan Schelten, Alex Vaughan, et~al. 2024.
\newblock The llama 3 herd of models.
\newblock \emph{arXiv preprint arXiv:2407.21783}.

\bibitem[{Huang et~al.(2024)Huang, Chen, Mishra, Zheng, Yu, Song, and Zhou}]{huang2024largelanguagemodelsselfcorrect}
Jie Huang, Xinyun Chen, Swaroop Mishra, Huaixiu~Steven Zheng, Adams~Wei Yu, Xinying Song, and Denny Zhou. 2024.
\newblock \href {https://arxiv.org/abs/2310.01798} {Large language models cannot self-correct reasoning yet}.
\newblock \emph{Preprint}, arXiv:2310.01798.

\bibitem[{Hurst et~al.(2024)Hurst, Lerer, Goucher, Perelman, Ramesh, Clark, Ostrow, Welihinda, Hayes, Radford et~al.}]{hurst2024gpt}
Aaron Hurst, Adam Lerer, Adam~P Goucher, Adam Perelman, Aditya Ramesh, Aidan Clark, AJ~Ostrow, Akila Welihinda, Alan Hayes, Alec Radford, et~al. 2024.
\newblock Gpt-4o system card.
\newblock \emph{arXiv preprint arXiv:2410.21276}.

\bibitem[{Kaushik et~al.(2020)Kaushik, Hovy, and Lipton}]{kaushik2020learningdifferencemakesdifference}
Divyansh Kaushik, Eduard Hovy, and Zachary~C. Lipton. 2020.
\newblock \href {https://arxiv.org/abs/1909.12434} {Learning the difference that makes a difference with counterfactually-augmented data}.
\newblock \emph{Preprint}, arXiv:1909.12434.

\bibitem[{Maas et~al.(2011)Maas, Daly, Pham, Huang, Ng, and Potts}]{maas-etal-2011-learning}
Andrew~L. Maas, Raymond~E. Daly, Peter~T. Pham, Dan Huang, Andrew~Y. Ng, and Christopher Potts. 2011.
\newblock \href {https://aclanthology.org/P11-1015} {Learning word vectors for sentiment analysis}.
\newblock In \emph{Proceedings of the 49th Annual Meeting of the Association for Computational Linguistics: Human Language Technologies}, pages 142--150, Portland, Oregon, USA. Association for Computational Linguistics.

\bibitem[{Madaan et~al.(2023)Madaan, Tandon, Gupta, Hallinan, Gao, Wiegreffe, Alon, Dziri, Prabhumoye, Yang, Gupta, Majumder, Hermann, Welleck, Yazdanbakhsh, and Clark}]{madaan2023selfrefineiterativerefinementselffeedback}
Aman Madaan, Niket Tandon, Prakhar Gupta, Skyler Hallinan, Luyu Gao, Sarah Wiegreffe, Uri Alon, Nouha Dziri, Shrimai Prabhumoye, Yiming Yang, Shashank Gupta, Bodhisattwa~Prasad Majumder, Katherine Hermann, Sean Welleck, Amir Yazdanbakhsh, and Peter Clark. 2023.
\newblock \href {https://arxiv.org/abs/2303.17651} {Self-refine: Iterative refinement with self-feedback}.
\newblock \emph{Preprint}, arXiv:2303.17651.

\bibitem[{Madaan et~al.(2021)Madaan, Padhi, Panwar, and Saha}]{madaan2021generatecounterfactualscontrolledcounterfactual}
Nishtha Madaan, Inkit Padhi, Naveen Panwar, and Diptikalyan Saha. 2021.
\newblock \href {https://arxiv.org/abs/2012.04698} {Generate your counterfactuals: Towards controlled counterfactual generation for text}.
\newblock \emph{Preprint}, arXiv:2012.04698.

\bibitem[{Neumann et~al.(2021)Neumann, Margolis, and Stanchi}]{neumann2021legal}
R.K. Neumann, E.~Margolis, and K.M. Stanchi. 2021.
\newblock \href {https://books.google.com.tw/books?id=ygMcEAAAQBAJ} {\emph{Legal Reasoning and Legal Writing}}.
\newblock Aspen Coursebook Series. Wolters Kluwer.

\bibitem[{Nguyen et~al.(2024)Nguyen, Youssef, Schlötterer, and Seifert}]{nguyen2024llmsgeneratingevaluatingcounterfactuals}
Van~Bach Nguyen, Paul Youssef, Jörg Schlötterer, and Christin Seifert. 2024.
\newblock \href {https://arxiv.org/abs/2405.00722} {Llms for generating and evaluating counterfactuals: A comprehensive study}.
\newblock \emph{Preprint}, arXiv:2405.00722.

\bibitem[{Ouyang et~al.(2022)Ouyang, Wu, Jiang, Almeida, Wainwright, Mishkin, Zhang, Agarwal, Slama, Ray et~al.}]{ouyang2022training}
Long Ouyang, Jeffrey Wu, Xu~Jiang, Diogo Almeida, Carroll Wainwright, Pamela Mishkin, Chong Zhang, Sandhini Agarwal, Katarina Slama, Alex Ray, et~al. 2022.
\newblock Training language models to follow instructions with human feedback.
\newblock \emph{Advances in neural information processing systems}, 35:27730--27744.

\bibitem[{Paranjape et~al.(2022)Paranjape, Lamm, and Tenney}]{paranjape-etal-2022-retrieval}
Bhargavi Paranjape, Matthew Lamm, and Ian Tenney. 2022.
\newblock \href {https://doi.org/10.18653/v1/2022.acl-long.117} {Retrieval-guided counterfactual generation for {QA}}.
\newblock In \emph{Proceedings of the 60th Annual Meeting of the Association for Computational Linguistics (Volume 1: Long Papers)}, pages 1670--1686, Dublin, Ireland. Association for Computational Linguistics.

\bibitem[{Shinn et~al.(2023)Shinn, Cassano, Berman, Gopinath, Narasimhan, and Yao}]{shinn2023reflexionlanguageagentsverbal}
Noah Shinn, Federico Cassano, Edward Berman, Ashwin Gopinath, Karthik Narasimhan, and Shunyu Yao. 2023.
\newblock \href {https://arxiv.org/abs/2303.11366} {Reflexion: Language agents with verbal reinforcement learning}.
\newblock \emph{Preprint}, arXiv:2303.11366.

\bibitem[{Williams et~al.(2018)Williams, Nangia, and Bowman}]{williams-etal-2018-broad}
Adina Williams, Nikita Nangia, and Samuel Bowman. 2018.
\newblock \href {https://doi.org/10.18653/v1/N18-1101} {A broad-coverage challenge corpus for sentence understanding through inference}.
\newblock In \emph{Proceedings of the 2018 Conference of the North {A}merican Chapter of the Association for Computational Linguistics: Human Language Technologies, Volume 1 (Long Papers)}, pages 1112--1122, New Orleans, Louisiana. Association for Computational Linguistics.

\bibitem[{Xu et~al.(2024)Xu, Zhu, Zhao, Pan, Li, and Wang}]{xu-etal-2024-pride}
Wenda Xu, Guanglei Zhu, Xuandong Zhao, Liangming Pan, Lei Li, and William Wang. 2024.
\newblock \href {https://doi.org/10.18653/v1/2024.acl-long.826} {Pride and prejudice: {LLM} amplifies self-bias in self-refinement}.
\newblock In \emph{Proceedings of the 62nd Annual Meeting of the Association for Computational Linguistics (Volume 1: Long Papers)}, pages 15474--15492, Bangkok, Thailand. Association for Computational Linguistics.

\bibitem[{Zhang et~al.(2015)Zhang, Zhao, and LeCun}]{Zhang2015CharacterlevelCN}
Xiang Zhang, Junbo~Jake Zhao, and Yann LeCun. 2015.
\newblock Character-level convolutional networks for text classification.
\newblock In \emph{NIPS}.

\end{thebibliography}

%\appendix

%\section{Example Appendix}
%\label{sec:appendix}

%This is an appendix.

\end{document}